%% file: arxiv_version.tex
 \documentclass[conference,a4paper,compsoc]{IEEEtran}
\usepackage{color}
\usepackage{graphicx}
\usepackage{amsmath}

\usepackage{amsfonts}

\usepackage{subfig}

\usepackage{float}

\ifCLASSINFOpdf
\else
\fi
\begin{document}
\title{Generalised Zero-Shot Learning with Domain Classification in a Joint Semantic and Visual Space}

% Authors at the same institution
\author{Rafael Felix$^{1,2}$, Ben Harwood$^{1,3}$, Michele Sasdelli$^{1,2}$ and Gustavo Carneiro$^{1,2}$ \\
Australian Centre for Robotic Vision$^{1}$\\
The University of Adelaide$^{2}$\\
Monash University$^{3}$\\
{\tt\small \{rafael.felixalves,michele.sasdelli,gustavo.carneiro\}@adelaide.edu.au, ben.harwood@monash.edu}
}

\maketitle

%\IEEEtitleabstractindextext{
\begin{abstract}
Generalised zero-shot learning (GZSL) is a classification problem where the learning stage relies on a set of seen visual classes and the inference stage aims to identify both the seen visual classes and a new set of unseen visual classes. Critically, both the learning and inference stages can leverage a semantic representation that is available for the seen and unseen classes.
Most state-of-the-art GZSL approaches rely on a mapping between latent visual and semantic spaces without considering if a particular sample belongs to the set of seen or unseen classes.
In this paper, we propose a novel GZSL method that learns a joint latent representation that combines both visual and semantic information. This mitigates the need for learning a mapping between the two spaces. Our method also introduces a domain classification that estimates whether a sample belongs to a seen or an unseen class. Our classifier then combines a class discriminator with this domain classifier with the goal of reducing the natural bias that GZSL approaches have toward the seen classes.
Experiments show that our method achieves state-of-the-art results in terms of harmonic mean, the area under the seen and unseen curve and unseen classification accuracy on public GZSL benchmark data sets. Our code will be available upon acceptance of this paper.
\end{abstract}
%}

% \IEEEdisplaynontitleabstractindextext
\IEEEpeerreviewmaketitle

%%%%%%%%%%%%%%%%%%%%%%%%%%%%%%%%%%%%%%%%%%%%%%%%
% INTRODUCTION
%%%%%%%%%%%%%%%%%%%%%%%%%%%%%%%%%%%%%%%%%%%%%%%%

%\IEEEraisesectionheading{
\section{Introduction}\label{sec:introduction}
%}

Humans have a powerful ability to learn about new visual objects without actually seeing them. This process generally involves the use of language to describe how a new visual object would look like. The textual description then allows for a new class of object to be formed in a person's mind. Our understanding of exactly how the human brain functions for this task is limited, but it is clear that humans make some sort of association between visual objects and semantic textual descriptions. Conceptually, objects with similar descriptions can naturally be viewed as being near to each other in some latent space, representing visual and semantic information. The research topic is known as generalised zero-shot learning (GZSL) aims to mimic this recognition ability of humans. In general, GZSL approaches employ an auxiliary set of semantic information that describes a set of visual classes. This additional information, such as tags or descriptions, can be utilised to overcome missing visual information in some of the classes~\cite{xian2017zero}.

\begin{figure*} %[Ht!]
	\centering
\includegraphics[width=17cm]{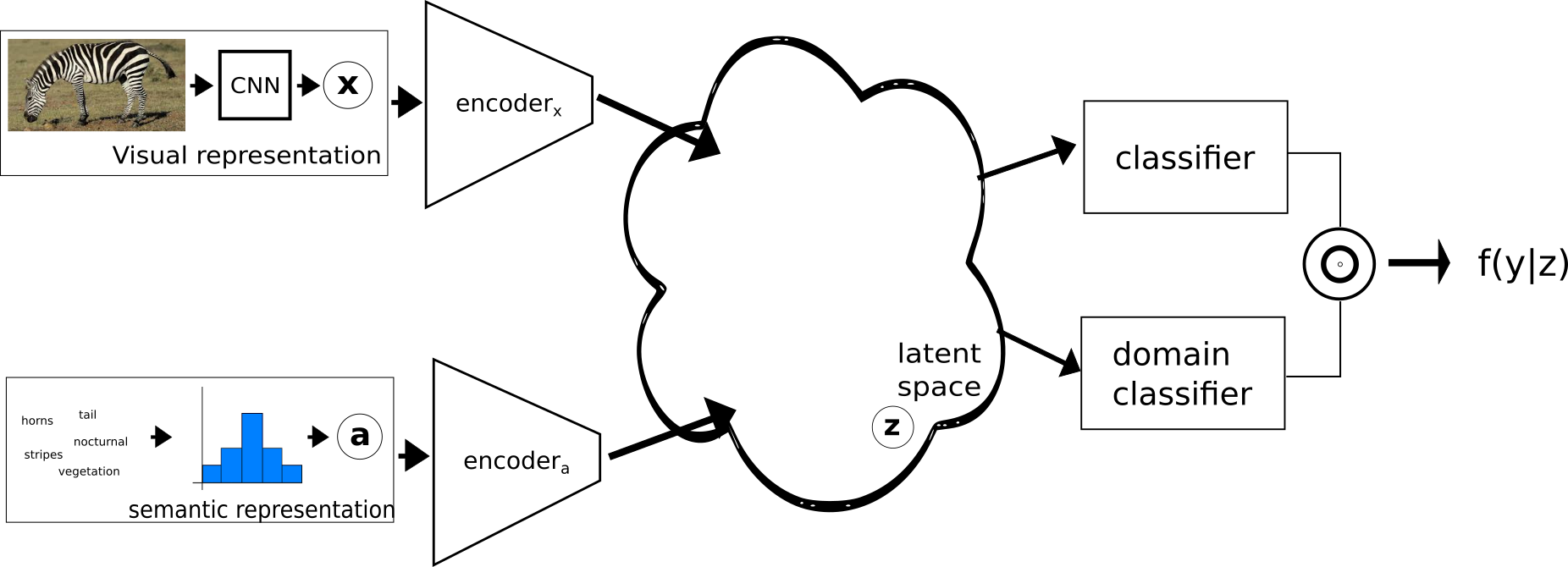}
    \caption{Depiction of the method proposed in this paper -- our approach learns the latent space for the visual and semantic modalities.  We train two classifiers using samples from this latent space: one to classify all the seen and unseen visual classes, and another to classify between the seen and unseen domains. The final classification combines the results of these two classifiers.}
	\label{fig:motivation}
\end{figure*}

Traditional GZSL approaches aim to recognise the visual classes available during the training process (i.e. the \textbf{seen}, source or known classes), and also classes that are not available during training (i.e. \textbf{unseen}, target or novel classes). Due to this constraint, GZSL approaches are intrinsically divided into two main tasks: (1) the training of a model that learns a transformation from the visual to the semantic space, using the visual samples and semantic information from seen classes; and (2) the transformation of a new test image by the model above into the semantic space, followed by a search of the closest semantic sample representing a seen or unseen class. In recent years, GZSL researchers have become increasingly interested in pairwise functions for disentangling these domains~\cite{frome2013devise}, and deep generative models~\cite{felix2018multi,schonfeld2019cada} for learning to transform between the visual and semantic representations. In general, GZSL methods do not try to estimate if a test sample belongs to the set of seen or unseen classes -- this issue inevitably biases GZSL approaches toward seen classes.  Only recently this issue has been acknowledged with a method that automatically combines the classification of Zero-Shot Learning (ZSL) for unseen classes with the classification of seen classes, by automatically weighting (using the test sample) the contribution of each classifier~\cite{atzmon2019adaptive}. Although that approach is in the right direction, it has the issue of relying on the training of multiple classifiers.
Another issue with the methods above is that they do not consider a latent space jointly optimised for the visual and semantic representation, which we believe is a crucial part of the inference process performed by humans that should be imitated by GZSL methods. In Fig.~\ref{fig:motivation}, we illustrate the idea explored in this paper for GZSL. The visual and semantic samples are represented in a joint latent space. This space is used to learn a classifier of visual classes and a domain classifier for seen and unseen domains.

In this paper, we aim to explore two observations about the latent space for the domain classification. The first observation is that samples from unseen classes that are visually similar to one of the seen classes tend to be projected relatively close to other seen classes distributions, instead of outside of the distribution of seen classes, as proposed by Socher et. al ~\cite{socher2013zero}. Our second observation is that samples from unseen classes that are visually different from any of the seen classes, tend to be projected outside the distribution of seen classes~\cite{socher2013zero}. Atzmon and Chechik~\cite{atzmon2019adaptive} propose a general framework that combines domain expert classifiers, such as DAP~\cite{lampert2009lerning} for unseen classes, and LAGO for the seen classes~\cite{atzmon2019adaptive}. However, this method relies on the disjoint training of both experts models, and the assumption that unseen samples are projected outside the distribution of seen classes~\cite{socher2013zero}. Hence, this method can be considered to be in general sub-optimal. We propose a general framework for learning and combining the visual and domain classifiers using the latent space. More specifically, we first introduce a general framework for latent space learning from cycle-WGAN~\cite{felix2018multi} and CADA-VAE~\cite{schonfeld2019cada}. Then, we propose a novel method for the seen and unseen domain classification from this latent space. Finally, we introduce a way to combine the visual and domain classifiers.
The empirical results show that our proposed framework outperforms previous approaches in terms of unseen accuracy and harmonic mean (\textbf{H-mean}) on several GZSL benchmark data sets, such as CUB~\cite{welinder2010caltech}, SUN~\cite{xian2017zero}, AWA1~\cite{lampert2009lerning,xian2017zero} and AWA2~\cite{lampert2009lerning,xian2017zero}. In terms of unseen accuracy, our method shows improvements of $4.5$\%, $5.6$\%, $2.5$\%, $1.5$\% for CUB, SUN, AWA1, and AWA2, respectively. Moreover, our method shows substantial improvements in terms of area under the curve of seen and unseen accuracy (\textbf{AUSUC}) \cite{chao2016empirical}. For AUSUC we improved from $0.3698$, $0.5238$, $0.5216$ to $0.3743$, $0.5247$, $0.5219$, on CUB, AWA1 and AWA2, respectively.

%%%%%%%%%%%%%%%%%%%%%%%%%%%%%%%%%%%%%%%%%%%%%%%%
% RELATED WORK
%%%%%%%%%%%%%%%%%%%%%%%%%%%%%%%%%%%%%%%%%%%%%%%%
\section{Related Work}
\label{sec:related}
% \section{Literature Review}

In this section, we discuss relevant literature that motivates and contextualises our work.

\subsection{Traditional Zero-Shot Learning}

Zero-shot learning (ZSL) is similar to GZSL, with a crucial difference: during inference, only the visual samples from the unseen classes are considered~\cite{lampert2014attribute,xian2017zero}. This difference makes ZSL a special case of GZSL. Therefore, critical problems present in GZSL are not considered in this approach, such as the natural bias of the visual classifier toward the seen classes.  Unfortunately, this setup not only reduces the applicability of ZSL methods but also makes it unrealistic for real-world applications~\cite{xian2017feature,felix2018multi}. Also, ZSL fails to handle jointly the seen and unseen data~\cite{chao2016empirical,changpinyo2016synthesized}. Due to the simplicity and unrealistic assumptions of ZSL, the whole field moved toward the GZSL problem, which is introduced in the next section.

\subsection{Generalised Zero-Shot Learning}

In GZSL, the algorithm is trained using visual samples from the seen classes, but the inference involves the analysis of samples from the seen and unseen classes. 
The main issue faced by GZSL methods is the bias toward the seen classes naturally present during inference, so a great deal of research has focused on mitigating this problem~\cite{felix2018multi,xian2017feature}. Particularly important examples of this type of research are anomaly detection~\cite{socher2013zero}, domain balancing~\cite{chao2016empirical} and generative data augmentation for GZSL~\cite{felix2018multi,schonfeld2019cada,xian2017feature}. Despite the advances in GZSL with the approaches mentioned above, we note that little attention has been devoted to addressing the seen/unseen domain classification in GZSL based on a latent space that is jointly learned to represent the visual and semantic representations. Moreover, we argue that the multi-modal nature of this joint latent space carries interesting properties to perform domain classification. In this paper, we show that classifying the seen and unseen domains plays an important role in improving domain balancing in GZSL.

\subsection{Data Augmentation for Zero-Shot Learning}

A particularly successful GZSL method is based on data augmentation, where artificial visual samples of the unseen classes are generated from the semantic representation to train the visual classifier~\cite{felix2018multi,schonfeld2019cada,xian2017feature,verma2018generalized}.  This approach has produced the current state-of-the-art results in GZSL benchmark data sets. Overall, these studies focus on how to learn generative models conditioned on the semantic information that is used to augment the data set for the unseen classes. Among the main approaches, we observe the use of Generative Adversarial Networks (GAN)~\cite{felix2018multi,xian2017feature} and Variational Autoencoders (VAE)~\cite{schonfeld2019cada,verma2018generalized}. In this paper, we formalise these approaches as a framework for generative probabilistic latent space learning.  Additionally, we show that these latent spaces have interesting properties that allow our approach to classifying samples into the seen or unseen domains for GZSL.

\subsection{Domain Classification}

Recent research has tackled the problem of GZSL as a novelty detection problem~\cite{socher2013zero}. This approach assumes that unseen classes are projected out of the distribution of seen classes. Therefore, these unseen classes samples can be handled as an outlier of the seen classes distribution~\cite{socher2013zero}. However, this approach fails to notice that samples from unseen classes can be projected relatively close to one of the seen classes. Atzmon and Chechik~\cite{atzmon2019adaptive} aims to tackle this novelty detection issue by providing a framework that handles domain classification for GZSL. The gist of that approach consists of a gating method that performs domain adaptation to combine an unseen class classifier (e.g.,  DAP~\cite{lampert2014attribute}, DeVISE~\cite{frome2013devise}), and CMT~\cite{socher2013zero}), and a seen class classifier~\cite{atzmon2019adaptive}.  Even though this method achieves remarkable performance in GZSL, it still relies on a sub-optimal disjoint training of multiple classifiers. In this paper, we mitigate these two issues by combining a seen/unseen class discriminator with a domain classifier that uses samples from a latent space that is trained to represent both the visual and semantic spaces.

%%%%%%%%%%%%%%%%%%%%%%%%%%%%%%%%%%%%%%%%%%%%%%%%
% METHOD
%%%%%%%%%%%%%%%%%%%%%%%%%%%%%%%%%%%%%%%%%%%%%%%%

\section{Method}\label{sec:method}

In this section, we introduce the problem formulation and our proposed approach.

\subsection{Generalised Zero-Shot Learning}\label{sec:problem_formulation}

In order to formulate the method of learning a classifier that can recognise visual samples from unseen visual classes, we define a visual data set
 $\mathcal{D} = \{(\textbf{x},  y)_i\}_{i=1}^{N}$
, where $\textbf{x} \in \mathcal{X} \subseteq \mathbb{R}^K$
 denotes the visual representation,
 and $y \in \mathcal{Y} = \{ 1,..., C \}$ denotes the visual class.
 Recent research shows that such visual representation, $\textbf{x}$, can be acquired from networks specialised in feature extraction.
 These are widely available in the literature, such as pre-trained deep residual nets~\cite{he2016resnet}.

In GZSL, the set of classes $\mathcal{Y}$ is split into two  domains: seen domain $\mathcal{Y}^S = \{1, ... , |S|\}$, and the unseen domain $\mathcal{Y}^U = \{(|S|+1), ... , (|S|+|U|)\}$. Hence, the total number of classes is $C=|S|+|U|$, with $\mathcal{Y} = \mathcal{Y}^S \cup \mathcal{Y}^U$, $\mathcal{Y}^S \cap \mathcal{Y}^U = \emptyset$. 
During training, we can only access visual samples from $\mathcal{Y}^S$, but during testing, samples can come from any class in $\mathcal{Y}$.
This lack of visual samples from unseen classes during training is compensated with a semantic data set that includes semantic information for the seen and unseen classes. Therefore, we introduce the semantic data set $\mathcal{R} = \{ (\mathbf{a},y)_j \}_{j \in \mathcal{Y}}$, which associates visual classes with semantic samples, where $\mathbf{a} \in \mathcal{A} \subseteq \mathbb R^L$ represents a semantic feature (e.g., set of continuous features such as \textit{word2vec}~\cite{xian2017zero}, or \textit{BoW}). 
Note that the semantic data set only has a single element per class. 

In comparison with the supervised learning paradigm, the problem of GZSL has a distinct setup. The data set $\mathcal{D}$ is divided into mutually exclusive training and testing visual subsets $\mathcal{D}^{Tr}$ and $\mathcal{D}^{Te}$, respectively. The $\mathcal{D}^{Tr}$ contains a subset of the visual samples belonging to the seen classes, and $\mathcal{D}^{Te}$ contains the visual samples from the seen classes that are held out from training and all samples from the unseen classes.
The training data set is composed of the semantic data set $\mathcal{R}$ and the training visual subset $\mathcal{D}^{Tr}$, while the testing data set relies only on the testing visual subset $\mathcal{D}^{Te}$.

\subsection{Data Augmentation Framework}

In this section, we first introduce the components for the latent space learning applied to GZSL models, then we describe CADA-VAE and cycle-WGAN. Finally, we introduce the domain classification for these latent space.

In recent years, we note an increasing number of models that use data augmentation for GZSL models \cite{felix2018multi,schonfeld2019cada,xian2017feature,kodirov2017semantic,mishra2018generative,zhu2018generative}. Overall, these methods aim to learn a generative model that produces artificial samples from unseen visual classes conditioned on their semantic representation. These artificial samples lie in a latent space.  In this paper, we aim to demonstrate that our proposed domain classification can be adapted to GZSL models that rely on data augmentation, such as CADA-VAE~\cite{schonfeld2019cada} and cycle-WGAN~\cite{felix2018multi}. Although these two models consist of different training approaches, we observe that their components can be generally described as a framework for latent space learning. Below, we introduce three components of such models: the encoder (or generator), the decoder (or regressor), and the discriminator.

The encoder transforms samples from an input space (i.e., visual or semantic) into a latent space. We represent the encoder with
\begin{equation}
\label{eq:encoder}
    \textbf{z}_x = Encoder_x(\textbf{x})
\end{equation}
for the visual space and similarly for the semantic space with $\textbf{z}_a = Encoder_a(\textbf{a})$, where the vector $\textbf{z}_{\{x,z\}} \in \mathbb R^Z$ lies in the latent space. The decoder transforms from the latent space into one of the input modalities. We represent the decoder with
\begin{equation}
    \label{eq:decoder}
    \tilde{\textbf{x}} = Decoder_x(\textbf{z}),
\end{equation}
and similarly for the semantic space with $\tilde{\textbf{a}} = Decoder_a(\textbf{z})$.  The latent space discriminator, used to determine whether a sample $\mathbf{z}$ belongs to the latent space given the input $\mathbf{x}$, is represented by
\begin{equation}
\label{eq:discriminator}
    p(\textbf{z} \mid \mathbf{x}) = Discriminator(\textbf{z};\mathbf{x}).
\end{equation}
We consider the simplified models above to describe CADA-VAE~\cite{schonfeld2019cada} and cycle-WGAN~\cite{felix2018multi} as the latent space learning models.

\begin{figure}[h!]
    \centering
    \includegraphics[width=3.5in,height=2in]{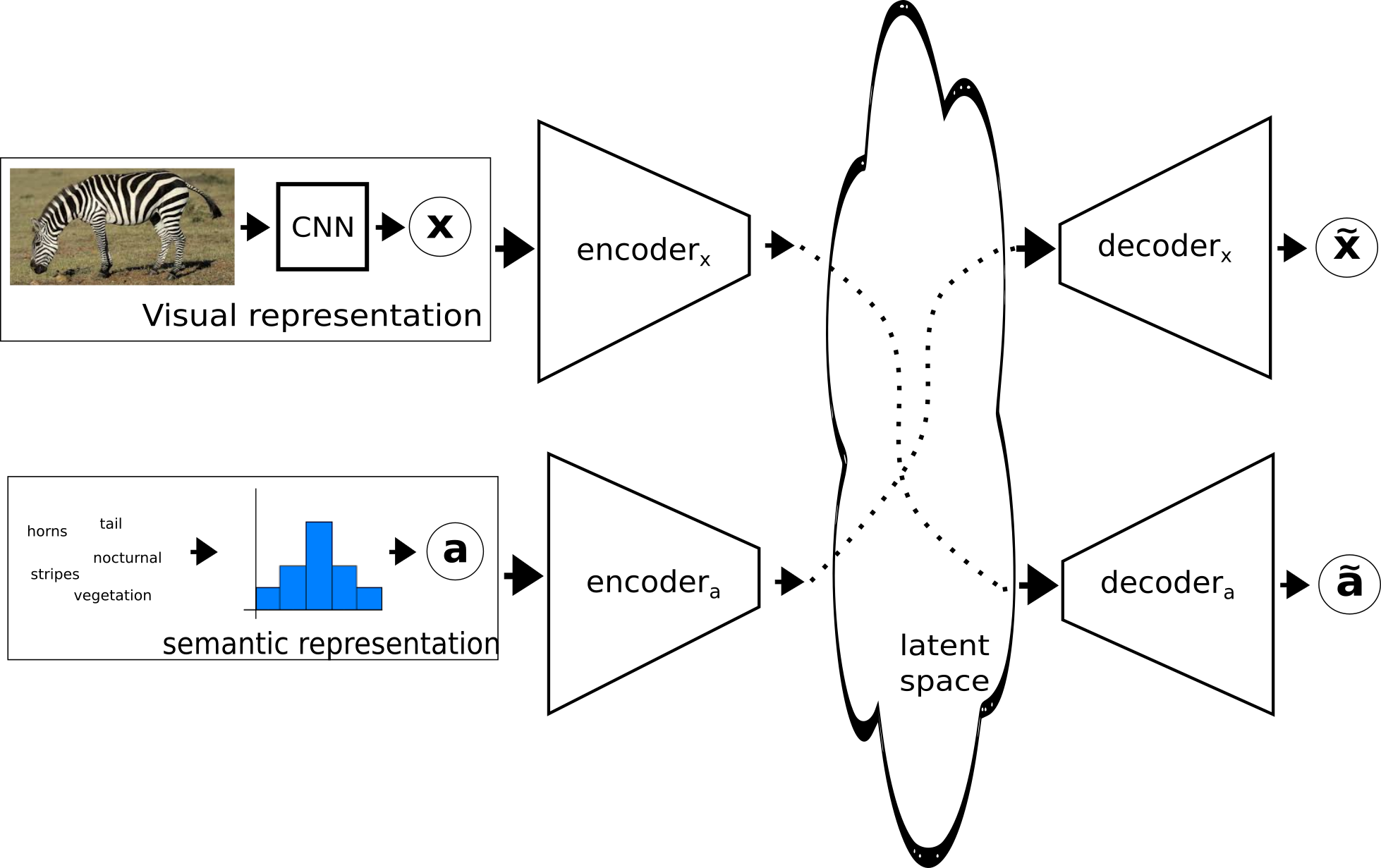}
    \caption{Depiction of the method CADA-VAE~\cite{schonfeld2019cada}. In this method encoders for the visual and semantic representation project samples into a shared latent space.}
    \label{fig:cada_vae}
\end{figure}

\textbf{CADA-VAE:} This model is a special type of variational autoencoder (VAE) for GZSL~\cite{schonfeld2019cada}. In this approach, the VAE aims to learn the latent space with cross alignment and distribution alignment losses, as depicted in Fig.~\ref{fig:cada_vae}. The overall loss by Schonfeld et al.~\cite{schonfeld2019cada} can be described with
\begin{equation}
\begin{split}
    \mathcal{L} = & \mathcal{L}_{VAE} 
                + \gamma \big(\sum_i^L \sum_{j \neq i}^L \mid\mid \textbf{x}^{(j)} - \tilde{\textbf{x}}^{(i)}\mid\mid \big)\\
                & + \delta \big(\mid\mid \mu^{(j)} - \mu^{(i)}\mid\mid_2^2 
                + \mid\mid \Sigma_{(j)}^{\frac{1}{2}} - \Sigma_{(i)}^{\frac{1}{2}}\mid\mid_{Frobenius}^2
                \big),
\end{split}
\label{eq:cada_vae}
\end{equation}
where the first term represents the VAE loss~\cite{schonfeld2019cada}, the second term denotes the reconstruction error between $L$ modalities -- that is, during training, the encoder projects input samples in the latent space (e.g. $Encoder_x$ for $\mathbf{x}$), then the decoder of a different modality is used (e.g.$Decoder_a$ from $\mathbf{z}_x$ -- see Fig.~\ref{fig:cada_vae}), which constraints the visual and semantic projections to be in the same region of the latent space represented by the mean $\mu$ and variance $\Sigma$ of the samples produced by the encoder~\cite{schonfeld2019cada}.  

\begin{figure}[h!]
    \centering
\includegraphics[width=3.5in,height=2in]{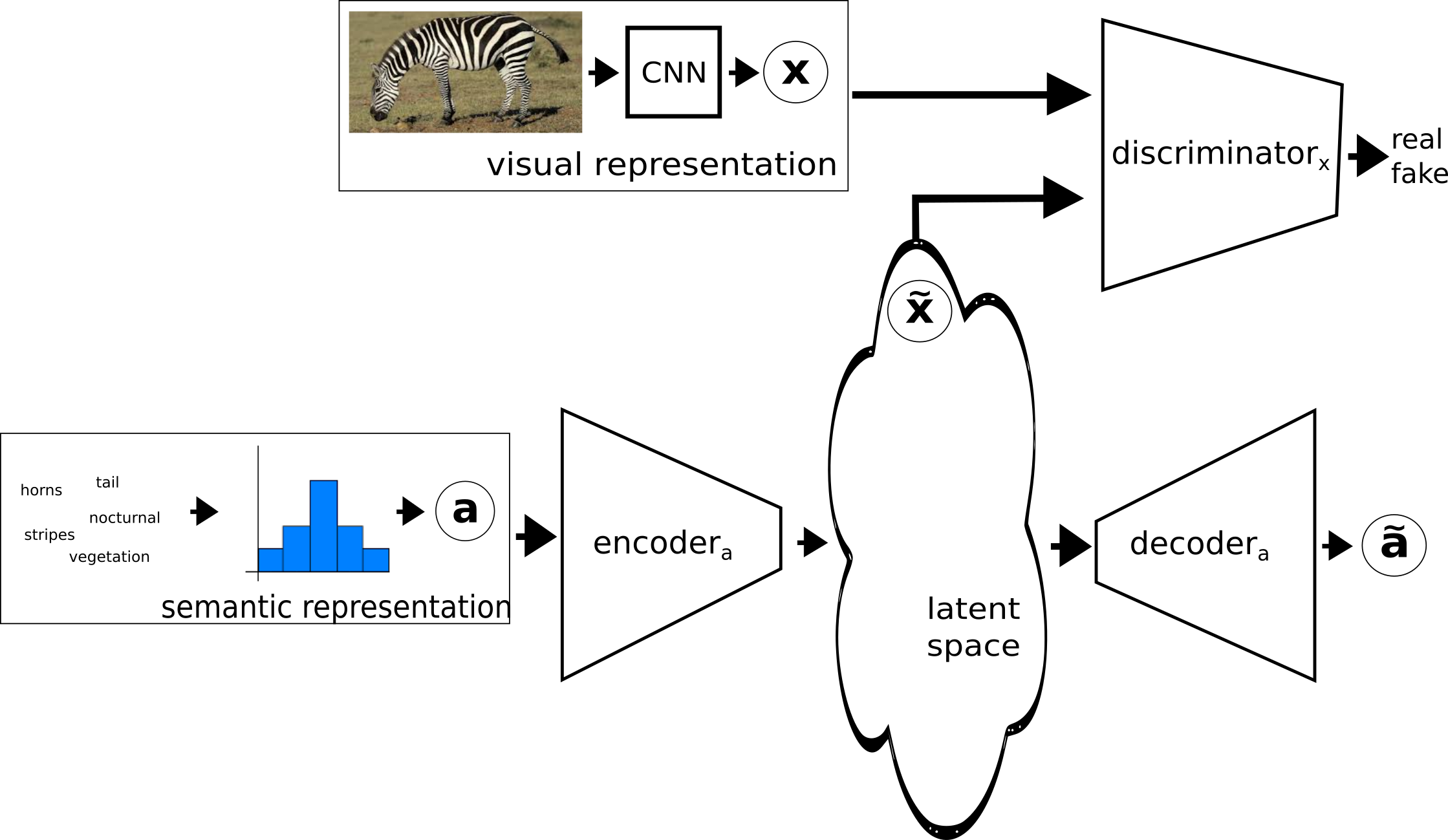}
    \caption{Depiction of the cycle-WGAN method~\cite{felix2018multi}. This method encodes the semantic space into a latent visual space. The decoder produces semantic vectors that are used to regularise the learning process.}
    \label{fig:cycle_wgan}
\end{figure}

\textbf{cycle-WGAN}: Fig.~\ref{fig:cycle_wgan} depicts the model cycle-WGAN~\cite{felix2018multi}. This model is optimised as a Generative Adversarial Network (GAN), regularised by a cycle consistent term, described with
\begin{equation}
    % \begin{split}
        \mathcal{L} = \mathcal{L}_{WGAN}
                    + \gamma \big( \mid\mid \textbf{a} - \tilde{\textbf{a}} \mid\mid_2^2 \big),  
    % \end{split}
    \label{eq:cycle_wgan}
\end{equation}
where the first term, $\mathcal{L}_{WGAN}$, represents a \textit{Wasserstein} Generative Adversarial Loss (WGAN~\cite{felix2018multi}), and the second term denotes the reconstruction loss (cycle) for the semantic representation. Thus, the generative projection of a given semantic representation into the latent space is encouraged to be back projected near the original semantic representation.

\subsection{Domain Classification}

From the previous section, we note that the latent space is an embedding space for visual and semantic samples. Therefore, we can use this latent space to learn a discriminative model given by
\begin{equation}
\label{eq:posterior_probability}
f( y \mid \mathbf{x}) = \int_{v} \int_{\mathbf{z}} p( \mathbf{z}\mid \mathbf{x}) f( y,v \mid \mathbf{z})  d v d\mathbf{z},
\end{equation}
where the function $f(.)$ represents the GZSL classifier and can be described in terms of domains, $v \in \{s,u\}$ ($s = seen$ and $u = unseen$), with
\begin{equation}
\label{eq:posterior_probability_sum}
f( y \mid \mathbf{x}) = \sum_{v\in \{s,u\}} p( y\mid \mathbf{z}_x) f( v \mid \mathbf{z}_x, y),
\end{equation}
where we assume from~\eqref{eq:posterior_probability} that $ p( \mathbf{z}\mid \mathbf{x})$ is a delta function at $\mathbf{z}_x = Encoder_x(\mathbf{x})$.
The term $p( y \mid \mathbf{z}_x)$ in \eqref{eq:posterior_probability_sum} is represented by a simple deep learning classifier with softmax activation. We define the function $f(.)$ in~\eqref{eq:posterior_probability_sum} by
\begin{equation}
    \begin{split}
    f( v \mid \mathbf{z}_x,y) = 
    &\begin{cases}
    p( v \mid \mathbf{z}_x, y), \text{~if~} v, y \text{ are in same domain}  \\
    0,\text{~otherwise,}
    \end{cases}
    \end{split}
    \label{eq:f_formalization}
\end{equation}
where "same domain" means the domain of seen or the unseen classes, and $p( v \mid \mathbf{z}_x, y)$ is denoted by a deep learning classifier with softmax activation.
The function in~\eqref{eq:f_formalization} represents our proposed domain classifier (DC). During the DC training, for training samples of the seen domain, we optimise $p( v = s | \mathbf{z}, y)$ with samples drawn from the latent space. These samples are acquired from visual and semantic representations projected in the latent space. For the unseen domain, $p( v = u | \mathbf{z}, y )$, we use the semantic projections in the latent space.

%%%%%%%%%%%%%%%%%%%%%%%%%%%%%%%%%%%%%%%%%%%%%%%%
% EXPERIMENTS
%%%%%%%%%%%%%%%%%%%%%%%%%%%%%%%%%%%%%%%%%%%%%%%%

\section{Experiments}
\label{sec:experiments}

In this section, we present the benchmark datasets, as well as the evaluation criteria for our experimental setup.  We then show the results of our method and compare them with the current state-of-the-art. Finally, we provide ablation studies to explore our method.

\subsection{Data Sets}
\label{sec:datasets}

\begin{figure}[h!]%
    \centering
    \subfloat{(A)}{{\includegraphics[width=1.4in,height=1.4in]{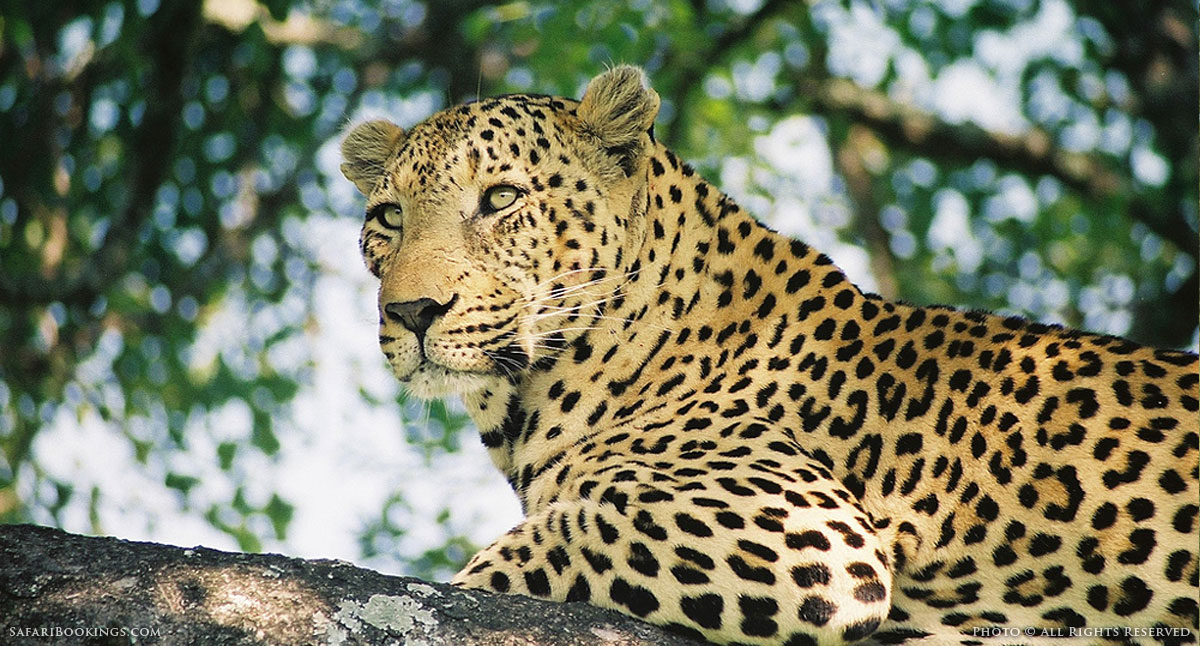} }}%
    \qquad
    \subfloat{(B)}{{\includegraphics[width=1.4in,height=1.4in]{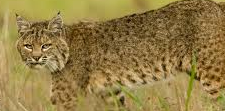} }}%
    \caption{Example of two classes that are visually similar from the benchmark dataset AWA1~\cite{xian2017zero}. (A) the sample leopard belongs to the seen classes, and (B) the sample bobcat belongs to the unseen classes. We speculate that samples from these two classes will lie close to each other in the latent space even though they come from different domains, challenging the view that samples from new unseen classes will lie far from samples of the seen classes in the latent space.}%
    \label{fig:example_visual_samples}%
\end{figure}

We assess our method on four publicly available benchmark GZSL data sets: CUB-200-2011~\cite{welinder2010caltech}; SUN~\cite{xian2017zero}; AWA1~\cite{lampert2009lerning,xian2017zero}, and AWA2~\cite{lampert2009lerning,xian2017zero}. To guarantee that our experiments are reproducible, we use the GZSL experimental setup described by Xian et al.~\cite{xian2017zero}. As the CUB data set is generally regarded as fine-grained, there is an intrinsic expectation that the novel unseen classes tend to have their class modes close to the seen classes. Thus, such dense visual representation space is a challenging problem for GZSL approaches. We also explore the use of coarse data sets, such as AWA1, AWA2, and SUN. Given the diversity of classes for such coarse data sets, there is an intrinsic expectation that novel classes will be projected far away from the samples of seen classes in the latent space, making the domain classification a trivial task. However, we argue that this statement does not always hold, particularly for classes that are visually similar (e.g. zebra/horse, whale/dolphin, leopard/bobcat), as depicted in Fig.~\ref{fig:example_visual_samples}. Table~\ref{table:dataset-stats} contains some basic information about the data sets in terms of the number of seen and unseen classes and the number of training and testing images.

\begin{table}[t!]
\centering
\caption{The benchmarks for GZSL: CUB\cite{welinder2010caltech}, SUN \cite{xiao2010sun}, AWA1\cite{xian2017zero}, and AWA2~\cite{xian2017zero}. Column (1) shows the number of seen classes, denoted by $|\mathcal{Y}^S|$, split into the number of training and validation classes (train+val), (2) presents the number of unseen classes $| \mathcal{Y}^U |$, (3) displays the number of samples available for training $|\mathcal{D}^{Tr}|$ and (4) shows number of testing samples that belong to the unseen classes $|\mathcal{D}_U^{Te}|$ and number of testing samples that belong to the seen classes $|\mathcal{D}_S^{Te}|$ from~\cite{felix2018multi,xian2017feature}}
\label{table:dataset-stats}
\begin{tabular}{|l|c|c|c|c|}
\hline
\textbf{Name} & $|\mathcal{Y}^S|$ (train+val) & $|\mathcal{Y}^U|$ & $|\mathcal{D}^{Tr}|$ & $|\mathcal{D}^{Te}_U|+|\mathcal{D}^{Te}_S|$
\\
\hline
\hline
CUB     & 150 (100+50) & 50 & 7057 & 1764+2967 \\
SUN    & 745 (580+65) & 72 & 14340 & 2580+1440\\
AWA1~\footnote{Dataset from https://cvml.ist.ac.at/AwA2/.}    & 40 (27+13) & 10 & 19832 & 4958+5685\\
AWA2 & 40 (27+13) & 10 & 23527 & 5882+7913\\
\hline
\end{tabular}
\end{table}

We represent the visual space by extracting image features from the activation of the 2048-dimensional top pooling layer of ResNet-101~\cite{he2016resnet}. For the semantic representation of the data set CUB-200-2011~\cite{xian2017zero}, we use the 1024-dimensional vector produced by CNN-RNN~\cite{reed2016learning}. These semantic samples represent a written description of each image using 10 sentences per image. To define a unique semantic sample per-class, we average the semantic samples of all images belonging to each class~\cite{xian2017zero}. We use manually annotated semantic samples containing 102 and 85 dimensions respectively, for the data sets SUN~\cite{xian2017zero}, AWA1~\cite{xian2017zero}, and AWA2~\cite{xian2017zero}. 
To prevent a violation of the ZSL constraints, where the test classes should not be accessed during training, all the features were extracted according to training splits proposed in~\cite{xian2017zero}.

\subsection{Evaluation Protocol}
\label{sec:evaluation_protocol}

Xian et. al ~\cite{xian2017zero} formalised the current evaluation protocol for GZSL. We first compute the average per-class top-1 accuracy measured independently for each class, then we calculate the overall mean. We calculate the mean-class accuracy for each domain separately, i.e., the seen ($\mathcal{Y}^S$) and the unseen ($\mathcal{Y}^U$) classes. Then, we also compute the harmonic mean (H-mean) of the seen and unseen domains accuracy~\cite{xian2017zero}. Furthermore, we show results by measuring the area under the seen and unseen curve (AUSUC)~\cite{chao2016empirical} by varying the domain expertise~\cite{chao2016empirical}. This domain expertise consists of a hyper-parameter to perform the trade-off between the performance in the seen and unseen classes~\cite{chao2016empirical}.

\subsection{Implementation Details}
\label{sec:implementation_details}

In this section, we describe the architecture and training procedures for learning the proposed latent space. As described in Sec.~\ref{sec:method}, we extend the following two models for our experimental setup: CADA-VAE~\cite{schonfeld2019cada} and cycle-WGAN~\cite{felix2018multi}. The model CADA-VAE contains the following models that are parameterised as neural networks: $Encoder_x(.)$, $Encoder_a(.)$ in~\eqref{eq:encoder}, $Decoder_x(.)$, and $Decoder_a(.)$ in \eqref{eq:decoder}. The training of CADA-VAE aims to produce a latent space that satisfies~\eqref{eq:cada_vae}. In terms of the model architecture and hyper-parameters (\textit{e.g. the number of epochs, batch size, the number of layers, learning rate, and, weight decay}), we followed the specifications provided by~\cite{schonfeld2019cada}. The encoder for visual representation is parameterised with $1560$ hidden neurons, and the encoder for the semantic representation is parameterised with 1450 hidden neurons. The decoders for the visual and semantic representation are parameterised with $1560, 660$  hidden neurons, respectively. For both modalities, the encoders project samples into the latent space, which is represented with 64-dimension vectors in the latent space. The model is optimised with Adam for 100 epochs~ \cite{kingma2014adam}. We use an adaptive scheduling rate for the hyper-parameters $\gamma, \delta$, by $(0.044, 0.0026)$, with respective epochs $(21-75, 0-90)$~\cite{schonfeld2019cada}.
We also extended cycle-WGAN \cite{felix2018multi}, as explained in Sec.~\ref{sec:method}. The model cycle-WGAN contains the following functions that are parameterised as neural networks: $Encoder_a(.)$ in \eqref{eq:encoder}, $Decoder_a(.)$ in \eqref{eq:decoder}, and $Discriminator(.)$ in \eqref{eq:discriminator}.  We followed the hyper-parameters choice (e.g. \textit{number of epochs, batch size, number of layers, learning rate, and weight decay, learning rate decay}) defined in~\cite{felix2018multi}. The encoder is parameterised with a single hidden layer containing 4096 nodes with LeakyReLU activation~\cite{maas2013relu}, and the output layer, with 2048 nodes, has a ReLU activation~\cite{nair2010rectified}. The decoder is parameterised with a linear layer, and the discriminator is a network with a single hidden layer with 4096 nodes. The network has a LeakyReLU activation, and the output layer has no activation.

The domain classifier (DC)\footnote{The code will be available upon acceptance, and we intend to add the link to the Github repository here.} is implemented as a neural network with binary output, representing the seen and unseen domains. The model is trained with Adam optimiser \cite{kingma2014adam} to recognise the domains. The output probability of the domain classifier tends not to be well calibrated~\cite{gal2017concrete,atzmon2019adaptive}. Therefore, we calibrate the model output using the validation set~\cite{gal2017concrete,xian2017zero}. Then, the domain classification is performed as described in~\eqref{eq:posterior_probability_sum}~\cite{chao2016empirical}.

\subsection{Results}
\label{sec:results}

In this section, we present the results for our proposed approach. The first question aimed to be answered in this paper consists of whether the proposed latent space contains relevant information that enables our approach to learn the domain classifier for GZSL. Thus, we provide numerical evidence that our method outperforms both baselines (i.e., CADA-VAE and cycle-WGAN) and previous GZSL. In Table~\ref{table:gzsl_results}, we show the results in terms of unseen class accuracy $\mathcal{Y}^U$, seen class accuracy $\mathcal{Y}^S$ and harmonic mean $H$, as described in Sec.~\ref{sec:evaluation_protocol}. These results are given for the data sets CUB, SUN, AWA1 and AWA2. We  compare  our approach with 12 leading GZSL methods, which are divided into three groups: semantic (SJE~\cite{akata2015evaluation}, ALE \cite{akata2016label}, LATEM~\cite{xian2016latent}, ESZSL~\cite{romera2015embarrassingly}, SYNC \cite{changpinyo2016synthesized}, DEVISE~\cite{frome2013devise}), latent space learning (SAE~\cite{kodirov2017semantic}, f-CLSWGAN \cite{xian2017feature}, {cycle-WGAN} \cite{felix2018multi} and CADA-VAE~\cite{schonfeld2019cada}) and domain classification (CMT\cite{socher2013zero} and DAZSL~\cite{atzmon2019adaptive}). The semantic group contains methods that only use the seen class visual and semantic samples to learn a transformation function from the visual to the semantic space, and classification is based on nearest neighbour classification in that semantic space. The latent space learning group relies on visual samples from seen classes and semantic samples from seen and unseen classes during training, and are detailed in Sec.~\ref{sec:method}.  The domain classification group relies on methods that weight the classification of seen and unseen classes. We discuss the numeral results in Table~\ref{table:gzsl_results} in  Section~\ref{sec:discussions}.

%%%%%%%%%%%%%%%%%%%%%%%%%
\input{src/gzsl-table.tex}
%%%%%%%%%%%%%%%%%%%%%%%%%

\subsection{Ablation Studies}
\label{sec:ablation}

In Table~\ref{table:gzsl_aucsuac} we report the area under the curve of seen and unseen accuracy (AUSUC)~\cite{chao2016empirical} for the benchmark data sets CUB, SUN, AWA1, and AWA2. We compare the results of the original CADA-VAE~\cite{schonfeld2019cada} and cycle-WGAN~\cite{felix2018multi} with and without the DC.
Similar to harmonic mean, the AUSUC is an evaluation metric that measures the trade-off between the seen and unseen domains.

\begin{table}[t!]
\centering
\caption{Area under the curve of seen and unseen accuracy (AUSUC).  The highlighted values per column represent the best results in each data set. The notation $\textbf{*}$ represents the results that we reproduced.}
\label{table:gzsl_aucsuac}

    \begin{tabular}{|l|c|c|c|c|}
        \hline
        \textbf{Classifier}  
        & \textbf{CUB} 
        % & \textbf{FLO} 
        & \textbf{SUN}
        & \textbf{AWA1 }
        & \textbf{AWA2 } \\
        \hline
        {EZSL}
        & $0.3020$ & $0.1280$ & $0.3980$ & $ -$ \\

        {DAZSL}~\cite{atzmon2019adaptive}
        & $0.3570$ & $\textbf{0.2390}$ & $0.5320$ & $-$ \\
        
        {f-CLSWGAN}~\cite{xian2017feature}
        & $0.3550$ & $0.2200$ & $0.4610$ & $-$ \\
        
        {cycle-WGAN}~\cite{felix2018multi}\textbf{*}~
        & $0.4180$ & $0.2321$ & $0.4730$ & $ - $\\

        {CADA-VAE}~\cite{schonfeld2019cada}\textbf{*}
        & $0.3698$ & $0.2362$ & $0.5238$ & $0.5216$ \\
        
        \hline
        \hline
        
        {cycle-WGAN + DC} 
        & $\textbf{0.4262}$ & $0.2321$ & $0.4744$ & $ - $\\
        {CADA + DC} 
        & $0.3743$ & $0.2364$ & $\textbf{0.5247}$ & $\textbf{0.5219}$ \\
        \hline
     
    \end{tabular}
\end{table}

%%%%%%%%%%%%%%%%%%%%%%%%%%%%%%%%%%%%%%%%%%%%%%%%
% DISCUSSIONS
%%%%%%%%%%%%%%%%%%%%%%%%%%%%%%%%%%%%%%%%%%%%%%%%

\section{Discussions}
\label{sec:discussions}

In this section, we discuss the main contributions presented by our approach. We performed our experiments by combining previous GZSL approaches (such as CADA-VAE~\cite{schonfeld2019cada}) and cycle-WGAN~\cite{felix2018multi}) with our Domain Classification in order to enhance the balancing of the seen and unseen domains for GZSL. 

Firstly, in Table~\ref{table:gzsl_results} we provide quantitative information that shows that our method outperforms existing methods in terms of unseen accuracy, $\mathcal{Y}^U$. This demonstrates that by learning to classify the domain for each sample, our method improves the classification of the unseen classes.  

Specifically, for CUB, SUN, AWA1 and AWA2 data sets, the baseline unseen classification results of $48.4$\%, $45.1$\%, $55.0$\%, and $55.5$\% have become $52.9$\%, $50.7$\%, $57.5$\%, and $57.0$\%. This improvement was achieved given a minor trade-off with the seen classes. 

Secondly, despite the trade-off mentioned above, our approach is still able to achieve minor improvements in terms of \textbf{H-mean}. Table~\ref{table:gzsl_results} shows an improvement of $0.2$\%, $0.1$\%, $0.1$\% and $0.4$\%, when compared to the baseline CADA-VAE. Although these results can be considered minor, we argue that our model does not directly optimise the H-mean. Thus, this improvement indicates that our approach has a more balanced performance than previous models.

We note similar behaviour for the cycle-WGAN model~\cite{felix2018multi}, where the proposed method achieves improvement for \textbf{H-mean} from $52.2$\% to $52.7$\% for CUB, from $39.2$\% to $40.3$\% for SUN, and from $59.7$\% to $60.0$\% for AWA1. However, such improvement is achieved due to the positive trade-off towards the seen domain. We argue that this difference, when compared to CADA-VAE, is due to the inherent differences in the latent space learning of each of the approaches. In fact, the approach CADA-VAE is directly optimised by a variational autoencoder,  where the control on the latent space is guided by a divergence measure for the visual and semantic representation jointly. On the other hand, the cycle-WGAN model is directly optimised by an adversarial loss from a generative adversarial network conditioned mainly on the semantic representation. 

In terms of AUSUC, the proposed approach achieves improvements for both cycle-WGAN~\cite{felix2018multi} and CADA-VAE~\cite{schonfeld2019cada}. For CADA-VAE, the domain classification yielded improvements from $0.3698$,$0.2362$, $0.5238$,$0.5216$ to $0.3743$, $0.2364$, $0.5247$, $0.5219$, for CUB, SUN, AWA1 and AWA2, respectively. Likewise, for cycle-WGAN~\cite{felix2018multi}, the DC provided improvements from $0.4180$, $0.4730$ to $0.4268$, $0.4744$ for CUB and AWA1, respectively.

%%%%%%%%%%%%%%%%%%%%%%%%%%%%%%%%%%%%%%%%%%%%%%%%
% CONCLUSIONS
%%%%%%%%%%%%%%%%%%%%%%%%%%%%%%%%%%%%%%%%%%%%%%%%
\section{Conclusion and Future Work}

In this paper, we introduce a principled method to classify the seen and unseen domains in GZSL. In particular, we presented our domain classifier that learns directly from the latent space of visual and semantic information. We have demonstrated that our proposed approach can be combined with previous latent space learning models, such as CADA-VAE and cycle-WGAN. Our approach yielded improvements for each one of those models by automatically balancing the seen and unseen domains in benchmark experiments on four available data sets: CUB, SUN, AWA1, and AWA2. 

Our experimental results show that our proposed approach has achieved state-of-the-art H-mean results for CUB, AWA1 and AWA2, and unseen accuracy for CUB, SUN, AWA1, and AWA2. In particular, our results are substantially better than the state of the art on CUB and SUN, which contain a large number of classes. On AWA1, AWA2, which are smaller data sets, our results are marginally better. Furthermore, our model produces substantial improvements in terms of AUSUC results for CUB, AWA1 and marginally better on AWA2.

As stated previously, our domain classification learns to discriminate between samples from the seen and unseen domains. We observe that the improvement of CADA-VAE and cycle-WGAN are different. The CADA-VAE model tends to improve in terms of the unseen domain when the DC is applied. Whereas cycle-WGAN tends to improve in terms of the seen domain. On one hand, we note that the training strategy for both models follows different guidelines, VAE and GAN. On the other hand, our model does not impose direct constraints in order to optimise GZSL metrics, such as accuracy or H-mean. In fact, we believe that these aspects are the main factors for the contrasting outcomes for CADA-VAE and cycle-WGAN models. With that in mind, we believe that the differences between these two data augmentation approaches should be studied in future generalised zero-shot learning research.

In the future, we intend to further study the reasons behind the performance difference observed between the data sets. Moreover, we also plan to develop a more extensive framework that can incorporate domain classification for approaches that do not rely on latent space learning.

\ifCLASSOPTIONcompsoc
  \section*{Acknowledgments}
\else
  \section*{Acknowledgment}
\fi

Supported by Australian Research Council through grants \textit{DP180103232}, \textit{CE140100016} and \textit{FL130100102}. We would like to acknowledge the donation of a TitanXp by Nvidia.

% \ifCLASSOPTIONcaptionsoff
%   \newpage
% \fi

\bibliographystyle{IEEEtran}
\bibliography{IEEEabrv,./bibliography.bib}

\end{document}

%% file: src/gzsl-table.tex
\begin{table*}[h!]
\centering
\caption{GZSL results using per-class average top-1 accuracy on the test sets of unseen classes $\mathcal{Y}^U$, seen classes $\mathcal{Y}^S$, and H-mean result $H$; and ZSL results on the unseen classes exclusively -- all results shown in percentage. The results from previously proposed methods in the field were extracted from~\cite{xian2017zero}. The highlighted values represent the best ones in each column. The methods below the double horizontal line represent the ones that use the semantic vectors from unseen classes during training. The notation $\textbf{*}$ represents the results that we reproduced, and results represented with $-$ were not available in the literature, or hyper-parameters were not given.}
\label{table:gzsl_results}
\centering

\resizebox{\textwidth}{!}{
\begin{tabular}{|l|lll|lll|lll|lll|}
\hline
& & \textbf{CUB} & 
% & & \textbf{FLO} &
& & \textbf{SUN} &
& & \textbf{AWA1} &
& & \textbf{AWA2} &
\\
\textbf{Classifier}  
&  $\mathcal{Y}^S$ & $\mathcal{Y}^U$ & $H$
&  $\mathcal{Y}^S$ & $\mathcal{Y}^U$ & $H$
&  $\mathcal{Y}^S$ & $\mathcal{Y}^U$ & $H$
&  $\mathcal{Y}^S$ & $\mathcal{Y}^U$ & $H$

\\ \hline \hline
\textbf{Semantic approach} &&& &&& &&& &&& \\

SJE~\cite{akata2015evaluation} &
$59.2$ & $23.5$ & $33.6$ & 
$30.5$ & $14.7$ & $19.8$ & 
$74.6$ & $11.3$ & $19.6$ & 
$73.9$ & $8.0$ & $14.4$ \\

ALE \cite{akata2016label} &
$62.8$ & $23.7$ & $34.4$ & 
$33.1$ & $21.8$ & $26.3$ & 
$76.1$ & $16.8$ & $27.5$ & 
$81.8$ & $14.0$ & $23.9$ \\

LATEM~\cite{xian2016latent} & 
$57.3$ & $15.2$ & $24.0$ & 
$28.8$ & $14.7$ & $19.5$ & 
$71.7$ & $7.3$ & $13.3$  & 
$77.3$ & $11.5$ & $20.0$ \\

ESZSL~\cite{romera2015embarrassingly} & 
$63.8$ & $12.6$ & $21.0$ & 
$27.9$ & $11.0$ & $15.8$ & 
$75.6$ & $6.6$ & $12.1$ & 
$77.8$ & $5.9$ & $11.0$ \\

SYNC \cite{changpinyo2016synthesized} &
$70.9$ & $11.5$ & $19.8$ & 
$43.3$ & $7.9$ & $13.4$ & 
$87.3$ & $8.9$ & $16.2$ & 
$90.5$ & $10.0$ & $18.0 $ \\

DEVISE~\cite{frome2013devise}  &
$53.0$ & $23.8$ & $32.8$ & 
$27.4$ & $16.9$ & $20.9$ & 
$68.7$ & $13.4$ & $22.4$ & 
$74.7$ & $17.1$ & $27.8 $ \\

\hline
\hline
\textbf{Generative approach} &&& &&& &&& &&& \\
SAE~\cite{kodirov2017semantic}        
& $18.0$ & $ 8.8$ & $11.8$
& $54.0$ & $ 7.8$ & $13.6$
& $77.1$ & $ 1.8$ & $ 3.5$
& $82.2$ & $ 1.1$ &	$ 2.2$
\\

f-CLSWGAN \cite{xian2017feature} & 
$57.7$ & $43.7$ & $49.7$ & 
$36.6$ & $42.6$ & $39.4$ & 
$61.4$ & $57.9$ & $59.6 $ & 
$68.9$ & $52.1 $ & $59.4$ \\

{cycle-WGAN} \cite{felix2018multi}
& $60.3$ & $46.0$ & $52.2$
% & $71.1$ & $59.1$ & $64.5$
& $33.1$ & $48.3$ & $39.2$
& $63.5$ & $56.4$ & $59.7$ 
& $  - $ & $  - $ & $  - $ \\ 

CADA-VAE~\cite{schonfeld2019cada} & 
$53.5$ & $51.6$ & $52.4$  & 
$35.7$ & $47.2$ & $40.6$  & 
$72.8$ & $57.3$ & $64.1$  & 
$75.0$ & $55.8$ & $63.9$ \\

{CADA-VAE}~\cite{schonfeld2019cada}\textbf{*} & 
$57.2$ & $48.4$ & $52.4$  & 
$36.8$ & $45.1$ & $40.6$  & 
$76.6$ & $55.0$ & $64.1$  & 
$75.3$ & $55.5$ & $63.9$ \\

\hline
\textbf{Domain Classification} &&& &&& &&& &&& \\
CMT \cite{socher2013zero} & 
$49.8$ & $7.2$ & $12.6$ & 
$21.8$ & $8.1$ & $11.8$ & 
$87.6$ & $0.9$ & $1.8$ & 
$90.0$ & $0.5$ & $1.0$ \\

DAZSL~\cite{atzmon2019adaptive}  &
$56.9$	& $47.6$ &	$51.8$ &
$37.2$	& $45.6$ &	$\mathbf{41.4}$ &
$76.9$	& $54.7$ &	$63.9$ &
$-$ & $-$ & $-$ \\

{cycle-WGAN} + DC (ours)
& $61.9$ & $45.9$ & $\textbf{52.7}$
& $39.3$ & $41.3$ & $40.3$
& $68.6$ & $53.4$ & $60.0$ 
& $  - $ & $  - $ & $  - $ \\ 

{CADA-VAE + DC} (ours)
& $52.4$ & $\textbf{52.9}$ & ${52.6}$
& $34.0$ & $\textbf{50.7}$ & $40.7$
& $72.6$ & $\textbf{57.5}$ & $\textbf{64.2}$
& $74.9$ & $\textbf{57.0}$ & $\textbf{64.3}$ \\

\hline

\end{tabular}
}
\end{table*}